# An Integrated Multi-Time-Scale Modeling for Solar Irradiance Forecasting Using Deep Learning

Sakshi Mishra*, Praveen Palanisamy*
*Microsoft Research + AI, Redmond, USA
Emails: sakshimishra@microsoft.com, praveen.palanisamy@microsoft.com

*Abstract*— **The amount of energy generation from renewable energy sources, particularly from wind and photovoltaic plants, has seen a rapid rise in the last decade. Reliable and economic operation of power systems thus requires an accurate estimate of the power generated from renewable generation plants, particularly those that are intermittent in nature. This has accentuated the need to find an efficient and scalable scheme for forecasting meteorological parameters, such as solar radiation, with better accuracy. For short-term solar irradiance forecasting, the traditional point forecasting methods are rendered less useful due to the non-stationary characteristic of solar power. The amount of operating reserves required to maintain reliable operation of the electric grid rises due to the variability of solar energy and the uncertainty of the forecasts. The ramp event caused by the variability of the solar resource magnifies the problem because the conventional large-generation plants cannot follow these rapid ramps. The higher the uncertainty in the generation, the greater the operating-reserve requirements, which translates to an increased cost of operation. In this research work, we propose a unified architecture for multi-time-scale predictions for intra-day solar irradiance forecasting using recurrent neural networks (RNN) and long-short-term memory networks (LSTMs). This paper also lays out a framework for extending this modeling approach to intra-hour forecasting horizons thus, making it a multi-time-horizon forecasting approach, capable of predicting intra-hour as well as intra-day solar irradiance. We develop an end-to-end pipeline to effectuate the proposed architecture. The performance of the prediction model is tested and validated by the methodical implementation. The robustness of the approach is demonstrated with case studies conducted for geographically scattered sites across the United States. The predictions demonstrate that our proposed unified architecture-based approach is effective for multi-time-scale solar forecasts and achieves a lower root-mean-square prediction error when benchmarked against the best-performing methods documented in the literature that use separate models for each time-scale during the day. Our proposed method results in a 71.5% reduction in the mean RMSE averaged across all the test sites compared to the ML-based best-performing method reported in the literature. Additionally, the proposed method enables multi-time-horizon forecasts with real-time inputs, which have a significant potential for practical industry applications in the evolving grid.**

## I. INTRODUCTION

The rise of renewable energy generation sources such as solar and wind, the expanded increase of energy storage systems, and the widespread adaption of demand response programs have changed the nature of the modern electric grid's operations. The intermittency of wind and solar generation is inherent, and there must be a mechanism to forecast the production from these resources with better accuracy to better schedule and manage power systems operations. Production forecasts of these generation technologies are tightly coupled with the forecasts of meteorological parameters, which increases the need for more accurate solar irradiance and wind forecasts. This work provides a scalable integrated modeling approach for very-short-term solar irradiance forecasting (nowcasting) using deep learning. This article is an extension of work originally presented in *IEEE Energy Conversion Congress and Exposition 2018* [1]. This article extends the work in [1] by implementing LSTM algorithm for predicting multi-time-scale solar irradiance using the proposed unified architecture, and comparing its performance with RNN algorithm. It also lays out a framework for broadening the scope of this modeling approach to intra-hour forecasting horizons using the proposed temporally-coupled unified architecture.

### A. *Motivation*

The electric grid is becoming increasingly dynamic in nature from the supply as well as the demand side. The main reasons for these changes in the grid are the increasing deployment of renewable energy generation sources and battery storage as well as expanded participation in demand response programs. Renewable energy penetration is growing due to its competitive levelized cost of energy (LCoE) leading to grid-parity and abundant availability. In order to ensure grid reliability and keep the operational costs sustainable, system operators need to gain real-time observability and obtain forecasts, with better accuracies, of intermittency of supply introduced by the renewable generation.

Uncertainty resulting from very-short-term forecasting errors has posed challenges for independent system operators (ISOs) in regions where the penetration of renewable energy generation is higher. In 2016, for managing the recurring errors in the short-term forecasting of renewable generation, the frequency regulation service requirements of the California ISO were doubled causing a sharp rise in the cost of frequency regulation.[1] This shows that the higher grid penetration levels of intermittent solar and wind generation pose a challenge for maintaining optimal grid reliability, consequently impacting the operational costs. However, accurate prediction of solar irradiance can mitigate this issue. A study conducted by Lew et al. [2] showed that by integrating wind and solar forecasts into its unit commitment (UC) process, the Western Electricity Coordinating Council (WECC) could save $5 billion per year.

Accurate prediction of solar irradiance plays a significant role in scheduling dispatches over various time-scales. For example, when the day-ahead solar forecast is incorporated in the UC process, the effectiveness of the UC gets boosted, resulting in decreased cost of operations. On the other hand, large forecast errors in very-short-term predictions caused by clouds passing over a large photovoltaic (PV) plant can induce voltage flickers, causing strain to the grid and leading to real-time frequency balancing issues. Similarly, other meteorological variables such as haze and dust storms due to wind speeds subject solar irradiance to sudden variation, triggering unacceptable voltage and frequency deviations that lead to grid instability.

For efficiently integrating ever larger amounts of solar energy with the grid, improvement in the accuracy of solar irradiance forecasts is a low-cost, high-impact solution and cannot be overemphasized. Thus, solar irradiance forecasting is a subject of paramount research because it plays a critical role in enabling high penetration of renewable generation on the electric grid globally. This research work presents an effective approach for very-short-term solar irradiance forecasting (i.e., nowcasting) for multi-time-scales (intra-hour and intra-day).

### *B. Literature Review*

Solar irradiance forecasting is a broad modeling problem that can be subdivided based on temporal and spatial/geographical resolutions for which the irradiance is being predicted. Various physics-based and statistical methods (including traditional machine learning methods) have been employed in the literature and are currently being used in the industry for different temporal and spatial resolution. A recent detailed review of solar-forecasting literature by Yang et al. [3] classifies solar-forecasting methods into five classes—namely, regression, numerical weather prediction, time series, image-based forecasting, and machine learning. Table 1 summarizes the methods, objectives, and applications of solar irradiance forecasting based on its temporal resolution.

The methods listed in Table 1 have been used for electric energy consumption (load) forecasting and renewable energy generation forecasting, with various degrees of predictive accuracy. The five main categories in which these methods can be divided are the following:

I. Statistical methods (based on regression methodologies), such as autoregressive (AR), AR integrated moving average (ARIMA), exponential smoothing (ES) models, and non-linear stationary models [4] [5] [6] [7].
II. Artificial intelligence (AI) methods, such as artificial neural networks (ANN) [8] [9] [10] [11] [12] [13], k-nearest neighbors [14], and fuzzy-logic systems (FLS) [15]
III. Physics-based models (numerical weather prediction [16] and persistence-based models [17])
IV. Sensing (remote and local) [18]
V. Hybrid models [19], e.g., satellite-derived cloud indices, and ANN [20] [21]and neuro-fuzzy systems [22].

Numerical weather prediction (NWP) models predict meteorological variables on the basis of physical laws of motion and thermodynamics that govern the weather. These models are powerful tools for predicting solar radiation for places where ground-based data are not available. However, the performance of these models is limited by their relatively coarse spatial resolution. The prediction of the extent and precise position of the cloud fields is challenging for these models. Thus, NWP models are expected to show inherent global and regional biases limiting forecasting accuracy. This is because the radiative properties of the clouds and the clouds themselves are difficult to incorporate in numerical models due to the limitation in spatial resolution of these model as well as the complex cloud microphysics. In order to overcome this shortcoming, NWPs are simulated at a regional level (termed Regional NWP models). Improved site-specific forecasts can be obtained by downscaling them to the regional level. Low temporal-resolution is another significant challenge for NWP models. For the Global Forecast System (GFS), the timescale of output variables of NWP models is from 3–6 hours and increases up to 1-hour for mesoscale models. However, applications such as predicting ramp-rate of the plant require intra-hour temporal-resolution, which limits the usefulness of current NWP models in such scenarios.

Satellite-based irradiance measurements have proven a useful tool for areas where ground-based measurements are not available [20]. The time evolution of the air mass is analyzed using images from a satellite, by superimposing images of the same area. Radiance is recorded by a radiometer installed in the satellite, and this radiance varies based on different atmospheric states (clear-sky to overcast). The main limitation associated with satellite sensing is the determination of an accurate set point for the radiance value under dense cloud conditions and under clear-sky conditions using each pixel in the image. Solar irradiance forecasting with remote sensing poses another limitation of using statistical and empirical algorithms [23] [24], which require ground-based solar data. The basis of these algorithms is a simple statistical regression between surface measurements and satellite information. Therefore, these algorithms do not employ precise information on the parameters that essentially model the solar radiation attenuation through the atmosphere.

[1] https://www.rtoinsider.com/caiso-regulation-requirements-29299/

**Table 1. Classification of methods used for forecasting generation and load for different time horizons**

| Forecasting Type | Time-Horizon | Objective | Applications | Methods | Inputs |
|---|---|---|---|---|---|
| Nowcasting | 5-min to 6 hours | Change in irradiance due to cloud motion and metrological variables changes | Economic dispatch, regulation, load-following, battery-use optimization, real-time market participation | - Statistical models with on-site measured irradiance (time-series forecasting)<br>- Machine Learning models (e.g., AVM, ARIMA)<br>- Image Processing models based on cloud motion vector from satellite images | - Meteorological Inputs (temperature, wind speed and direction, relative humidity, pressure, clearness index), both historical and predicted<br>-Time series Solar Irradiance measurements<br>-Sky Images |
| Short-term forecasting | 6-hours to 3 days | Change in irradiance due to daily weather variations | Unit commitment, switching source, or re-scheduling means of production | NWP models combined with statistical and machine learning algorithms | -Meteorological Inputs (temperature, wind speed and direction, relative humidity, pressure, clearness index), both historical and predicted<br>-Time series Solar Irradiance measurements |
| Medium-term forecasting | Up to several months | Captures seasonal variation patterns | Schedule maintenance, bidding in capacity markets | Statistical Models | Historical Solar Resource and Weather Data |
| Long-term forecasting | Several years | Captures climate change impacts on solar resource | Planning for large-scale PV plant deployment: Long-term purchase agreements, Forward Capacity Market | Statistical Models | Historical Solar Resource and Weather Data |

These limitations of the remote sensing models and NWP for nowcasting have steered the research towards using statistical models and more recently AI, including machine learning (ML) techniques for conducting the time-series analysis of the predicted variable (solar irradiance). As described in [25], statistical methods can mainly be classified as:

I. Linear stationary models (moving average models, autoregressive models, mixed autoregressive moving average models, and mixed autoregressive moving average models with exogenous variables)
II. Nonlinear stationary models
III. Linear non-stationary models (autoregressive integrated moving average models (ARIMA) with or without exogenous variables)

These classical statistical methods provide several advantages over NWP and sensing methods, such as higher temporal-resolution and eliminating the need for remotely sensed data. However, these individual statistical techniques are often limited by strict assumptions of linearity, variable independence, and normality. For solar irradiance forecasting, these methods often yield large errors in cloudy-sky conditions, due to the non-stationarity of the solar irradiance data series. Also, these methods do not provide a way to use a single model to predict solar irradiance for more than one time-step at a time. There is research work conducted in recent literature [26] to reconcile solar forecasts from various models (which are designed to predict for different time-scales) to produce the revised forecast for a given time-step with better accuracy than the base forecast.

Being universal function approximators, ANNs are capable of representing complex nonlinear behaviors in a high-frequency, non-stationary, and high-dimensional dataset such as solar irradiance. When exogenous variables such as pressure, temperature, and humidity are considered for modeling the global horizontal irradiance, ANNs model the complex nonlinear relationship between these exogenous input variables and map their relationship with the global horizontal irradiance (GHI, the output variable). Deep Learning (DL) is a subset of machine learning methods that learns the intricate structure in large datasets, as opposed to task-specific algorithms. This gives DL algorithms the power to generalize over previously unseen scenarios. An ANN with multiple hidden layers between input and output layers is called a Deep Neural Network (DNN), and DNNs have proven their effectiveness in solving complex problems in various fields including automatic speech recognition, natural language processing, and image recognition due to the advances in computational capabilities over the last decade [27].

Feedforward neural networks (FFNNs), with multiple hidden layers, have been extensively used for solar irradiance/load predictions in the literature [8] [9] [10] [11] [12]. However, FFNNs have the underlying assumption of independence among the time-series samples or the data points. Therefore, after processing each time-series sample, the entire state of the network is cleared, and the network starts mapping the input and output for the next time step from scratch, thereby not accounting for the impact of the previous timestep's exogenous variables on the next time-step's predicted variable.

Recurrent neural networks (RNNs) address this limitation of FFNNs [28]. RNNs are networks with a loop in them that helps with persisting or carrying over the information from previous time steps. In theory, RNNs are capable of handling long-term temporal dependencies, as well. However, in practice, RNNs do not seem to be able to learn them [29]. Long short-term memory networks – called "LSTMs" – are a special kind of RNN, capable of learning long-term dependencies. They were introduced by Hochreiter and Schmidhuber in 1997 [30]. LSTM networks are explicitly designed to overcome the long-term dependency problem encountered by their parent algorithm (RNN), and so, they can be applied for short-term solar irradiance forecasting problems, as well. RNNs have traditionally been difficult to train because of the millions of parameters present between various layers, requiring high computational power. This limitation has been overcome by the recent advances in network architectures, parallelism, graphics processing units (GPUs), and optimization techniques enabling successful large-scale learning.

Deep learning algorithms have recently been applied to photovoltaic power generation forecasting in the literature [31] [32]. The approach employed in [31] takes NWP predictions as inputs to the DL model and forecasts PV power generation. Essentially, the algorithm is designed to depend on the forecasts from NWP algorithms to output PV power generation, making it ineffective for intra-hour power generation forecasts. The problem of predicting daily aggregated solar radiation is addressed in [32]. The scope is limited to aggregated daily PV power generation forecasting in this work. Further, DL has been applied to solar irradiance forecasting in [33]. However, the problem is formulated as time-series forecasting without exogenous variable input. Moreover, with 10-milliseconds time resolution, the paper trains and tests the model on a total of four days. Solar irradiance forecasting as a time-series problem, with the hourly resolution, is evaluated in [34], as well. This study again does not consider any exogenous variables.

This research work attempts to fill the gap in applying deep learning to solar irradiance forecasting in a temporally-coupled architecture that takes past values of exogenous variables along with the past solar irradiance measurements as input to predict solar irradiance for multiple future time steps using a single model. The next section delineates the contributions of this work.

### *C. Our Contributions*

Solar forecasting is not a new research area, and there have been several useful methods proposed in the past. However, most of the methods documented in the literature predict the irradiance for a particular time horizon, and in the multi-time-scale context, no single model performed well in comparison with other models. Moreover, the state-of-the-art methods proposed for solar irradiance forecasting primarily make averaged rather than instantaneous forecasts (e.g., a minute-to-minute scale). This work proposes a novel unified architecture that fills the gap between the need for improvement in forecasting accuracy and the need for temporally integrated models. Our proposed method takes high-dimensional time-series meteorological data as input and makes predictions with a forward inference time on the order of milliseconds, enabling near real-time forecasts based on the measured data of the current time-step. This offers a great value for system operators who require real-time multi-time-scale forecasting for effectively managing frequency regulation, load-following, and economic dispatch with high penetration of renewables on the grid. The specific contributions of this research work are as follows:

- We demonstrate that deep learning architecture-based models can be used to effectively predict solar irradiance at very-short-term timescales (minutes to hours) based *only* on historical meteorological data.
- We propose two approaches for very-short-term solar irradiance predictions:
  - A single system with the capability to output solar forecasts for 1-hour, 2-hour, 3-hour, or 4-hour time-scales.
  - A temporally-coupled unified architecture that can forecast the solar irradiance for multiple time-scales; for example, the trained model is capable of forecasting the solar irradiance values for the 1-hour, 2-hour, 3-hour and 4-hour time horizons.
- We present extensive experimental evidence on large scale datasets for locations that are geographically apart to demonstrate the effectiveness of the proposed unified architecture in climatically unique weather situations that span the United States.
  - We experiment with deep learning models based on both Recurrent neural network (RNN) and Long-short term memory (LSTM) neural network structures and demonstrate that DL models outperform traditional ML models used to predict for a single time-scale (i.e., 1-hour or 2-hour) at a time.
- We propose a framework to extend this unified architecture to *multi-time-horizon* (minute-by-minute as well as hourly resolution) for very-short-term prediction applications. This framework enables intra-hour and intra-day predictions to be made using a unified architecture.

### D. *Organization of the paper*

With the motivation behind this research work laid out in the first section, along with the review of the literature and the list of contributions of this research work, the rest of the paper is organized as follows. Section II introduces the preliminaries of solar irradiance forecasting, and preliminaries of the DL approaches leveraged in this work. Section III details the proposed unified architecture based on RNN and LSTM, solidifies the problem statement and introduces the evaluation metric. Section IV presents the case study conducted for seven locations in the United States, the results of the case study, and a thorough analysis of the results. Section V summarizes the work and identifies future avenues of research in enhancing the proposed solar irradiance forecasters' effectiveness and widening the time-horizon covered by this forecaster using deep learning algorithms.

## II. PRELIMINARIES

This work broadly encompasses concepts from two fields: i) deep learning (branch of machine learning) and ii) solar forecasting. In the following subsections, we establish the preliminaries associated with these two field that are utilized in the proposed methodology.

### A. *Deep Learning*

#### 1) *Recurrent Neural Network*

The differentiating factor between FFNNs and RNNs is the additional directed edges present in RNNs. The notion of a temporal component is introduced into the model by these edges, which span adjacent time steps, essentially forming feedback connections in the hidden-layer nodes. Thus, RNNs are capable of processing sequences of inputs by using their internal memory, thereby exhibiting dynamic temporal behavior. In other words, RNNs have the form of a chain of repeating modules of neural nets that unfold over time. This property of RNNs can be harnessed in predicting the solar irradiance for very short-term forecasting by considering the measured irradiance for the previous time steps.

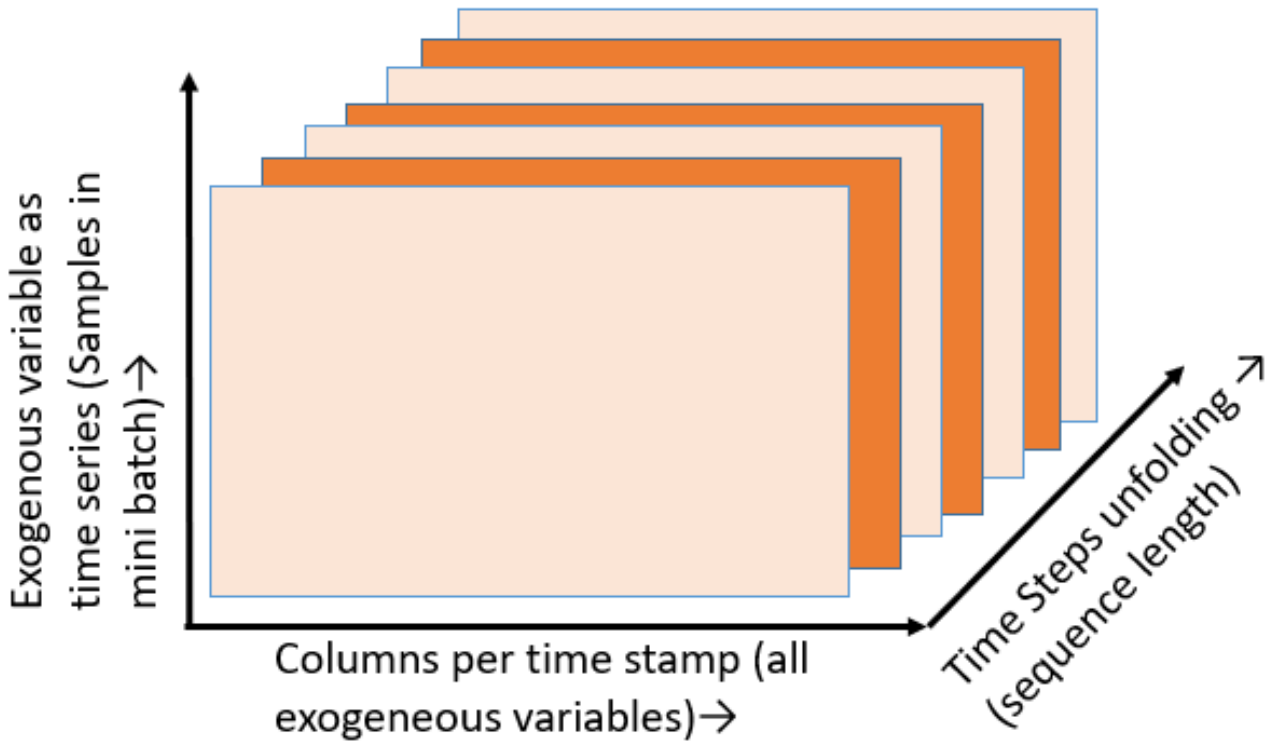

**Figure 1. Recurrent neural network input representation**

RNN takes input as a sequence, and its target can be a single value or a sequence. The input sequence is denoted by an array of the real-valued vector $(x^{(1)}, x^{(2)}, \dots, x^{(t)})$, where $x^{(t)}$ represents each data-point/sample at time-step t. $(y^{(1)}, y^{(2)}, \dots, y^{(t)})$ denotes the target array. Each datapoint $y^{(t)}$ can either be a scalar or a real-values vector. The predictions are denoted by $(\boldsymbol{\hat{y}}^{(1)}, \boldsymbol{\hat{y}}^{(2)}, \dots, \boldsymbol{\hat{y}}^{(3)})$. Thus, the input to the RNN has three dimensions (Figure 1):

I. Mini-batch size or the time series data-points (or more appropriately data-vectors for this specific problem formulation with exogenous variables)

II. Input feature vector or number of columns in the vector per time step

III. The number of time-steps or the sequence length; this dimension unfolds the input vector over time.

In an RNN network, nodes with recurrent edges receive input from the current input sample $x^{(t)}$and additionally receive hidden node values preserved from the previous time step $h^{(t-1)}$, at time-step $t$, as shown in equation (1). The hidden state value thus obtained from equation (1) is used in equation (2) for predicting the output $\boldsymbol{\hat{y}}^{(t)}$ at $t+n$, with $n$ being the prediction-horizon of interest.

$$\boldsymbol{h}(t) = f_h( \boldsymbol{W}_{hx} x^{(t)} + \boldsymbol{W}_{hh} h^{(t-1)} + b_h) \quad (1)$$

$$\boldsymbol{\hat{y}}^{(t)} = f_o(\boldsymbol{W}_{yh} h^{(t)} + b_y) \quad (2)$$

In the equations above, $f_h$ and $f_o$ are the activation functions of hidden and output layers, respectively. $\boldsymbol{W}_{hx}$ is the weight matrix between the input and the hidden layer (similar to a FFNN), and $\boldsymbol{W}_{hh}$ is the recurrent weights matrix between the hidden layer and itself at adjacent time-steps. $b_h$ and $b_y$ are bias parameters.

*2) Long Short Term Memory Network*

LSTM is a specific type of RNN, where the repeating module in the chain-like structure (as RNN) itself is a memory cell. This memory cell consists of four layers of neural networks embedded in each cell, interacting in a special way. The key difference between an RNN and LSTM is that RNN always replaces the content of a unit with a new value computed from the current input $x^{(t)}$ and the previous hidden state $h^{(t-1)}$ , whereas LSTM keeps the existing content and appends the unit with the incoming information.

The concept of cell states $(C_t)$ is an important differentiating factor of LSTM. The information flows through the cell state with updates to this information through carefully regulated structures called gates. Forget gate $(f_t)$, input gate $(i_t)$, and output gate $(o_t)$ are the three main gates, as shown in Figure 2, which facilitate the function of regulating the information update to the cell state. The forget gate layer takes in the previous hidden state $h^{(t-1)}$and the current input and passes it through the sigmoid function $x^{(t)}$. It outputs a number between 0 and 1 for each number in the cell state $C_{t-1}$ in equation (3), where 1 represents "do not remove any information from the previous cell state" and 0 represents "completely remove it." The input gate decides which of the new incoming information will be stored in the next state. There are two parts to this process. First is the input gate layer which decides which values will be updated, as shown in equation (4). Next, a vector of new candidate values, $\tilde{C}_t$, is created by a hyperbolic tangent (tanh) layer, which could be added to the state equation (5). The new cell state is updated from $C_{t-1}$to $C_t$ by adding the input gate's output and the new candidate values, equation (6). Now, the last gate (the output gate) functions as the filter for outputting the new hidden state $h^{(t)}$, as depicted in the Figure 2 and shown in equations (7) and (8).

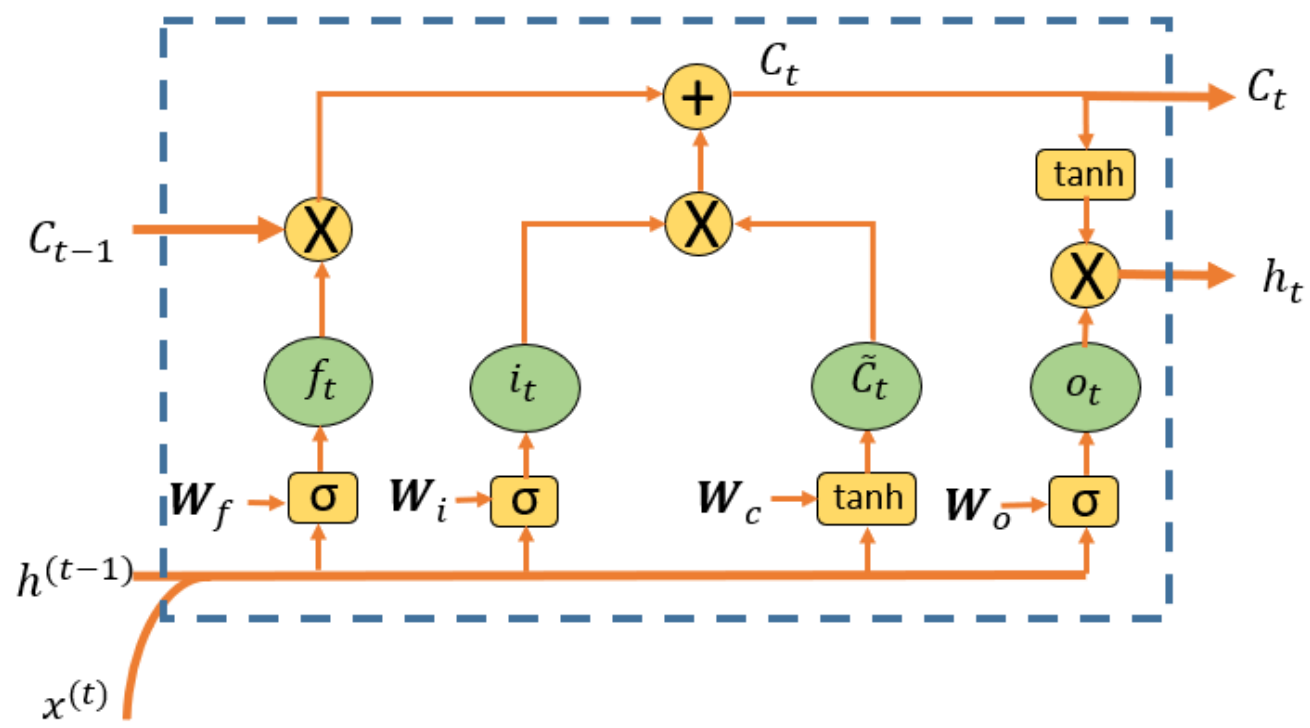

**Figure 2. LSTM memory cell diagram**

$$f_t = \sigma\,( W_f . [\, h^{(t-1)} , x^{(t)}] + b_f) \quad (3)$$

$$i_t = \sigma\,(W_i.[\,h^{(t-1)}, x^{(t)}] + b_i) \qquad (4)$$

$$\tilde{C}_t = tanh\,(W_c.[\,h^{(t-1)}, x^{(t)}] + b_c) \qquad (5)$$

$$C_t = f_t * C_{t-1} + i_t * \tilde{C}_t \qquad (6)$$

$$o_t = \sigma\,(W_o.[\,h^{(t-1)}, x^{(t)}] + b_o) \qquad (7)$$

$$h^{(t)} = o_t * \tanh\,(C_t) \qquad (8)$$

where $W_f$, $W_i$, $W_c$, $W_o$ are the weights and $b_f$, $b_i$, $b_c$, $b_o$ are the biases for forget, input, new candidate, and output layers, respectively [35].

*3) Deep Learning Algorithm*

Once the RNN (and, in extension, LSTM) architecture unfolds the input along the sequence dimension, as shown in Figure 3, the network then becomes analogous to the FFNN. This unrolled network is then trained using backpropagation through time algorithm (BPTT) [36] iteratively across the sequence dimension consisting of various time steps. The learning algorithm thus consists of two main iterative phases:

I. Forward propagation: The input vector is propagated forward through the layers (by multiplying it with the weight vector and passing it through the activation function) to obtain the output vector.

II. Weight update: The loss function is calculated by finding the error between the network's interim prediction output and the target. This error is then propagated backward to calculate the gradient for updating weight matrices.

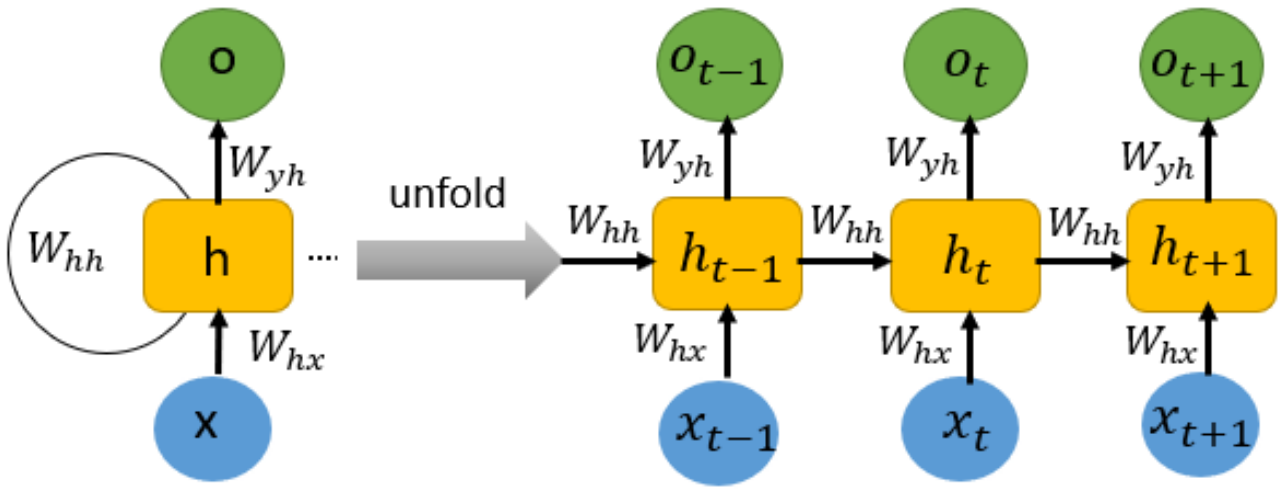

**Figure 3. RNN unfolding**

The "learning process" terminates depending on the preset threshold of the error amount or a preset number of epochs for iterating through the whole training dataset. In this way, the learning process helps the neural network to discern the trends in historical data (by internally modeling the complex nonlinear relationships) and provides highly accurate very-short-term predictions without requiring a specific mathematical model a priori.

### *B. Generative AI and Predictive AI – A Note*

Considering the on-going mainstream deployment of generative AI application, we make a clear distinction between generative AI and predictive AI; the latter one's application being explored in the presented work. Predictive AI systems, as the name indicates, are designed to predict the outcomes based on past data patterns and available knowledge. More specifically, supervised machine learning based predictive AI identifies the trends, correlations, and statistical patterns in datasets. Predictive AI's strength and salient property is its ability to analyze vast amounts of historical data and make highly accurate predictions and estimation about future events. Generative AI, on the other hand, uses an altogether distinct strategy which emphasizes the production of fresh and creative content (in the form of text, images, audio, video). The algorithms behind generative AI produce creative new results that resemble human inventiveness. In short, generative AI creates new content based on the learned patterns and predictive AI makes predictions about future events based on the historic data.

### *C. Solar*

*1) Solar Irradiance: From a Physics-Based Perspective*

*Irradiance*, expressed in Watts/m$^2$, is the rate at which radiant energy is incident on a surface per unit area of the surface. *Irradiation*, expressed in $\mathrm{J/m^2}$ or $\mathrm{kWh/m^2}$, is the incident energy per unit area on a surface, found by integration of irradiance over a specified time. The total solar irradiance ($G_{TOT}$) reaching the surface of the Earth has three components: 1) beam horizontal radiation ($G_{BHR}$); 2) diffuse solar irradiance ($G_{DIFF}$); and 3) the ground-reflected irradiance ($G_{REFL}$). On a horizontal surface at the level of the ground, the total irradiance is the sum of the direct and diffuse radiations (the reflected radiance is often neglected). $G_{BHR}$ , given by equation (10), varies with the orientation of the receiving surface whereas $G_{DIFF}$ is considered the same for all the surfaces. The total solar irradiance ($G_{TOT}$) for a terrestrial surface of any orientation and tilt with an incident angle θ is given by equation (9) [37].

$$G_{TOT} = G_{DNI}.\cos(\theta) + G_{DIFF}.\left(\frac{1+\cos(\beta)}{2}\right) + G_{REFL}.\left(\frac{1-\cos(\beta)}{2}\right) \quad (9)$$

in which,

$$G_{BHR} = A_0 e^{\frac{-B}{\sin(\beta)}} \quad (10)$$

where $A_0$ is the apparent extraterrestrial radiation at air mass $m = 0$; $B$ is the atmospheric extinction coefficient; and $\beta$ is the sun's altitude above the horizon, in degrees [38].

#### 2) *Clear-sky Model*

A knowledge of clear-sky conditions, characterized by the absence of clouds, is required for the calculation of the clear-sky index parameter $K_t$ (described later) in the proposed methodology. The clear-sky model used in this work, the Bird Clear-sky model, is described in [39]. Bird developed a transmittance expression for the different attenuation processes in the atmosphere based on a Radiative Transfer Model (RTM) calculation. The model takes three meteorological input parameters—the broadband aerosol optical depth, ozone column, and water vapor—and calculates direct and diffuse solar insolation $GHI_{clear}$.

This physics-based formulation of irradiance is very useful for understanding the physical underpinnings of the phenomenon. The clear-sky models are also used for persistence forecasts. However, predicting the irradiance for a future time is done through various other methods, as described in the literature review section.

## III. THE PROPOSED METHODOLOGY

Different from the physics-based statistical modeling approach, the proposed methodology for predicting the solar irradiance is centered around mapping the nonlinear functional relationships between the measured meteorological variables in the past and current time-steps and the solar irradiance in the future horizon of interest. Deep neural networks can model the complex relationships between the sequence of various exogenous variables (measured meteorological variables) and their combined impact on the predicted variable (solar irradiance) dynamically. RNNs and LSTMs, due to their specialized architecture, also incorporate temporal aspects in mapping these complex relationships.

### A. *Solar Irradiance Forecasting: A Supervised Learning Problem*

We formulate the solar irradiance forecasting as a supervised learning problem where the data is divided into training and testing sets. The clear-sky index values, $K_t$ (the prediction target, described later in this section) at time step $t + n$ (n=1,2,3,4 hours), are made available to the network during the training process using the training data set. During testing, the $K_t$ values are only used for reporting the error between the expected and actual outcome. The exogeneous meteorological input variables used are:

- Downwelling global solar ($\mathrm{Watts/m^2}$)
- Upwelling global solar ($\mathrm{Watts/m^2}$ )
- Direct-normal solar ($\mathrm{Watts/m^2}$)
- Downwelling diffuse solar ($\mathrm{Watts/m^2}$)
- Downwelling thermal infrared ($\mathrm{Watts/m^2}$),
- Downwelling infrared case temperature (K)
- Downwelling infrared dome temperature (K),
- Upwelling thermal infrared ($\mathrm{Watts/m^2}$)
- Upwelling infrared case temperature (K)
- Upwelling infrared dome temperature (K)
- Global UVB ($\mathrm{milliWatts/m^2}$)
- Photosynthetically active radiation ($\mathrm{Watts/m^2}$)
- Net solar (dw_solar – uw_solar) ($\mathrm{Watts/m^2}$)
- Net infrared (dw_ir – uw_ir) ($\mathrm{Watts/m^2}$)
- Net radiation (netsolar + netir) ($\mathrm{Watts/m^2}$)
- 10-meter air temperature (C)
- Relative humidity (%)

- Wind speed (m/s)
- Wind direction (Degrees, clockwise from north)
- Station pressure (mb).

The clear-sky index $K_t$, is defined as the ratio of the measured irradiance to the modeled clear-sky irradiance at ground level. At time t, clear-sky GHI, obtained using the Bird model, is denoted by $GHI_{clear}^{t}$. With the assumption of zero cloud coverage, this represents the theoretical GHI at time *t*. Clear-sky index is thus given by equation (11).

$$Kt_i^{(t)} = \frac{GHI^t}{GHI_{clear}^t} \qquad (11)$$

where $GHI^t$ is the instantaneously observed value. For hourly predictions, it is averaged over the forecasting horizon (f.h.). The averaged hourly clear-sky index ending at time f.h. is thus given by equation (12).

$$Kt_a^{(f.h.)} = \frac{\sum_{s=f.h.-60}^{f.h.} Kt_i^{(s)}}{60} \qquad (12)$$

*B. End-to-End Algorithmic Pipeline*

The pipeline is designed to ingest the given dataset with meteorological exogenous variables, also taking in the clear-sky component, and processing it through the core RNN and LSTM algorithm to predict the clear-sky index for the chosen forecasting horizon. The $GHI^t$ value can then be obtained from multiplying the predicted clear-sky index by $GHI_{clear}^t$. The flowchart of the proposed unified architecture for predicting solar irradiance for multiple time horizons using RNN and LSTM is shown in Figure 4. There are three main modules encapsulating the overall algorithm:

*1) Pre-processing*

The site-specific data consisting of the aforementioned meteorological variables is imported. The Bird Model [39] is used for obtaining the clear-sky global horizontal irradiance values for the specific sites. These two different datasets are merged based on the time-stamp. The dataset is then split into training and testing sets. Then the clear-sky index parameter is calculated. Being a ratio of two quantities with the same units, $K_t$ is a dimensionless parameter. It is a common practice to filter out the data in order to remove night hours and to objectively compare the studied predictors, concerning the global radiation forecasting. The insignificant value of solar irradiation present during these hours justifies this choice of elimination. The mean of the column and/or neighborhood values are used to fill in the missing data. In order to normalize the dataset, the extreme outliers needed to be identified and eliminated. Exploratory data analysis accomplishes the same.

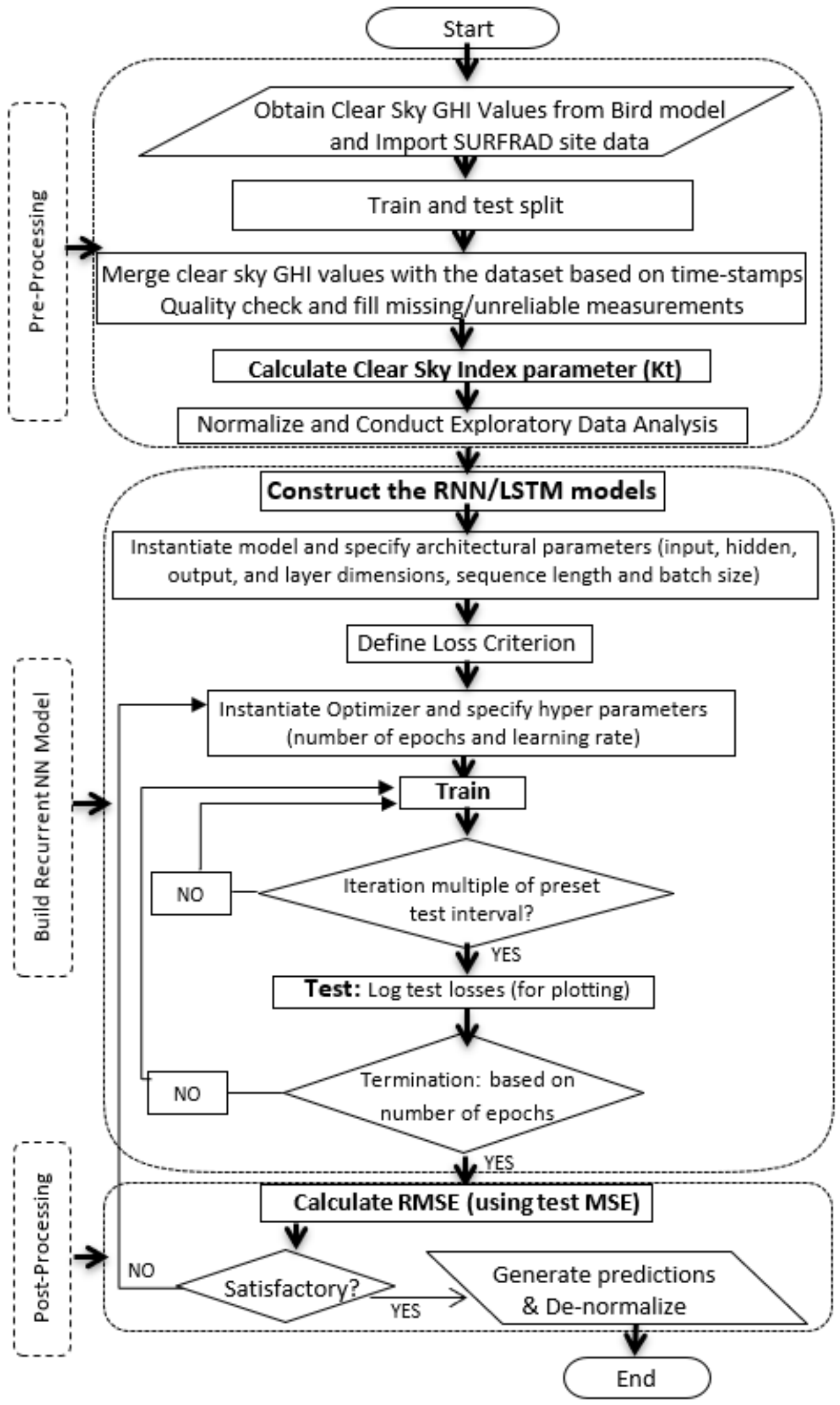

**Figure 4. Flow chart of the proposed unified architecture**

*2) RNN and LSTM training and testing*

RNN and LSTM model classes are constructed. The architectural parameters such as input dimension (number of nodes in the input layer), hidden dimension (number of nodes in the hidden layer), layer dimension (number of hidden layers), and output dimension (number of nodes in the output dimension) are initialized, and the models are instantiated. The output dimension for our proposed methodology is set to be equal to the number of time horizons for which the prediction are made, which is four for the case study conducted in this work. For the fixed time-horizon prediction, the architecture shown in Figure 5 (left) is used, which maps multiple input features to a single output variable. For the multi-time-horizon predictions, the architecture shown in Figure 5 (right) is used, which maps multiple input features to multiple output variables (many-to-many mapping).

The sequence dimension is also an architectural parameter that is specified during the model-instantiation stage. This parameter specifies the number of time-steps for which the input will be unrolled by the RNN and LSTM algorithms. Figure 1 shows the sequence dimension as the third dimension of the functional mapping of temporal aspect in the RNN and LSTM networks. The activation function used for the hidden layers in the RNN algorithm is rectified linear unit (ReLU) [40]. Since the solar irradiance prediction is a regression type problem, the output layers use linear units as their activation function. In the LSTM algorithm, tanh and sigmoid activation functions are used in different layers of the memory cell as depicted in Figure 2.

*3) Post-processing*

The post-processing stage involves the denormalization of the MSE. The RMSE values are calculated from the stored MSE. Hyperparameter tuning (including learning rate, number of epochs, number of hidden nodes, batch size) is then conducted for obtaining satisfactory RMSE. This tuning is again an iterative process where reasonable hyperparameters are obtained empirically. The training process is completed once satisfactory (and expected) RMSE values are obtained. The predicted clear-sky index can then be multiplied with the clear-sky GHI to obtain the forecast of the expected GHI in real time.

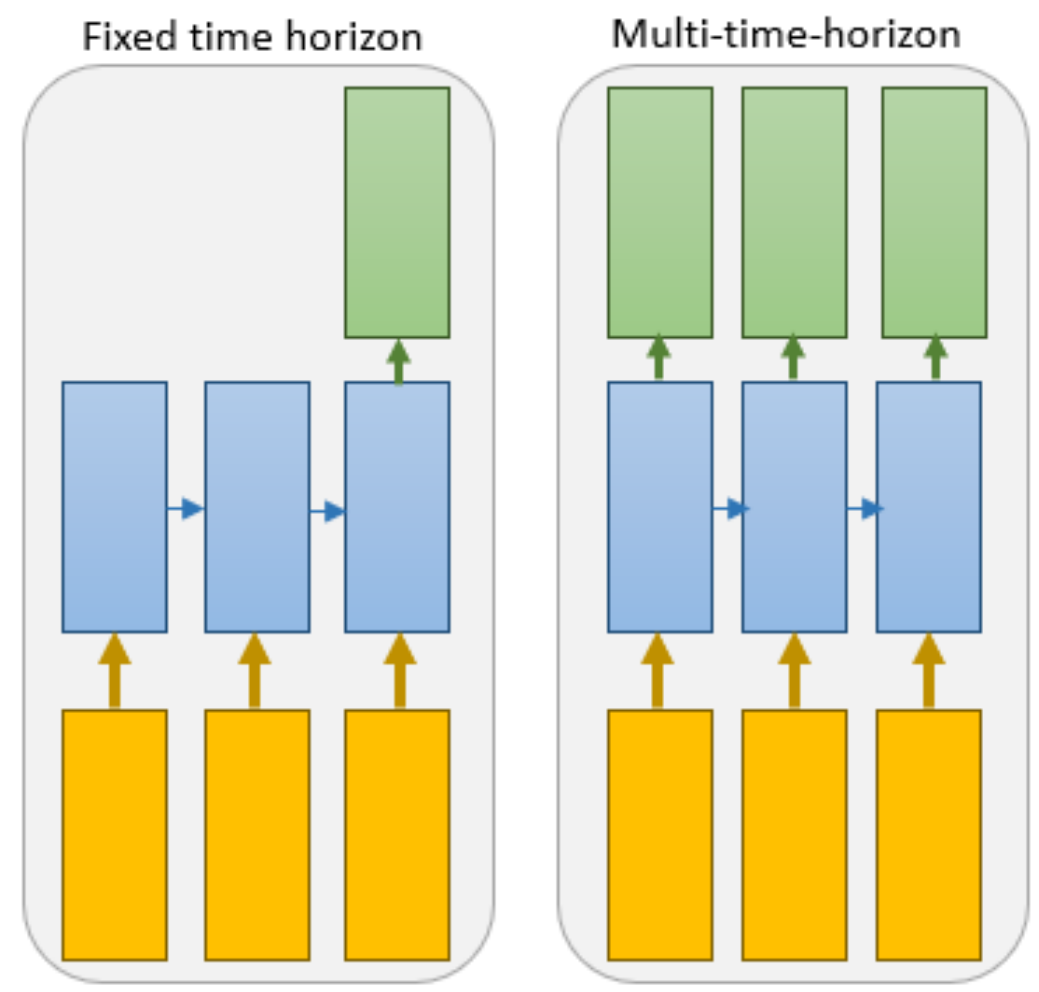

**Figure 5. Input-output mapping used in the proposed method**

### *C. Evaluation Metric*

Mean squared error (MSE) is used as a measure of accuracy by this algorithm to find the difference between the actual value (also called target value, denoted by T) and the output that the neural network produces during the training process (also called predicted value, denoted by P). The MSE formula is shown in equation (13). For the purpose of benchmarking, during the post-processing stage, the RMSE, shown in equation (14), is calculated by taking the square root of the MSE values.

$$MSE = \frac{1}{n}\sum_{i=1}^{n}\left(T_i^{(t)} - \widehat{P_i^{(t)}}\right)^2 \qquad (13)$$

$$RMSE = \sqrt{\frac{1}{n}\sum_{i=1}^{n}\left(T_i^{(t)} - \widehat{P_i^{(t)}}\right)^2} \qquad (14)$$

where $T_i^{(t)}$ is a vector of target values and $\widehat{P_i^{(t)}}$ is a vector of predicted values.

## IV. Case Study

Seven Surface Radiation Budget Network (SURFRAD) stations across the continental United States provide an accurate nationwide database[2] to study the effectiveness of the proposed solar irradiance forecaster. SURFRAD stations are equipped with Eppley precision spectral pyranometers (PSPs). These pyranometers are capable of measuring GHI to within ±2% [41]. NOAA's Solar Radiation Facility is used for calibrating PSPs. An average of sixty 1-second instantaneous measurements are reported as 1-minute GHI [42]. Night-time GHI values are generally negative due to pyranometer thermal offset. For this reason, all negative values of GHI were set to zero in this work.

[2] https://www.esrl.noaa.gov/gmd/grad/surfrad/

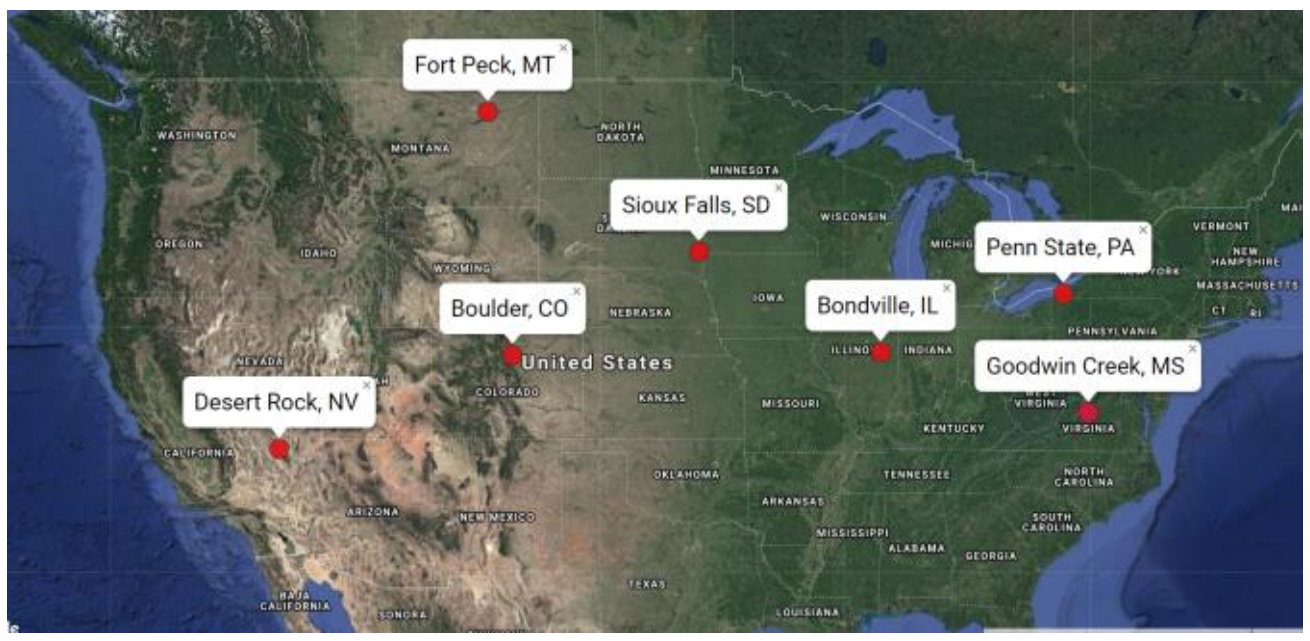

**Figure 6. The seven SURFRAD sites in the case study (ref map)**

Boulder, CO; Desert Rock, NV; Fort Peck, MT; Sioux Falls, SD; Bondville, IL; Goodwin Creek, MS; and Penn State, PA [Figure 6] are the seven SURFRAD sites considered in the case study. These sites are geographically spread across the United States, providing diverse climate profiles to test the robustness of the proposed forecasting approach. The dataset from the years 2010 and 2011 are used to train the algorithm. To benchmark against the results reported in the literature, the test year for each site is chosen to be 2009 for the single time-horizon forecasting approach. In this well-structured and decoupled train and test year experiment design, there is no arbitrary partition of the data into training and testing sets. There are seven different models trained for these seven sites. The approach is thus temporally-coupled (integrated forecasters for various time horizons) and spatially-decoupled (different forecasting models for different geographical locations).

### A. *Exploratory Data Analysis*

Observed downwelling solar irradiance and clear-sky global horizontal irradiance have a good correlation, as shown in Figure 7 and Figure 8. The values plotted in these figures are average daily solar radiation for a whole year (2017), at Boulder location. The clear-sky global horizontal irradiance, by definition, does not consider the cloud cover in calculating the irradiance. On the other hand, the observed global horizontal index is a measured quantity that is recording the actual irradiance for a given set of spatial coordinates. From the SURFRAD dataset, downwelling solar irradiance is chosen to be equivalent to the global horizontal index because of the high correlation between the two.

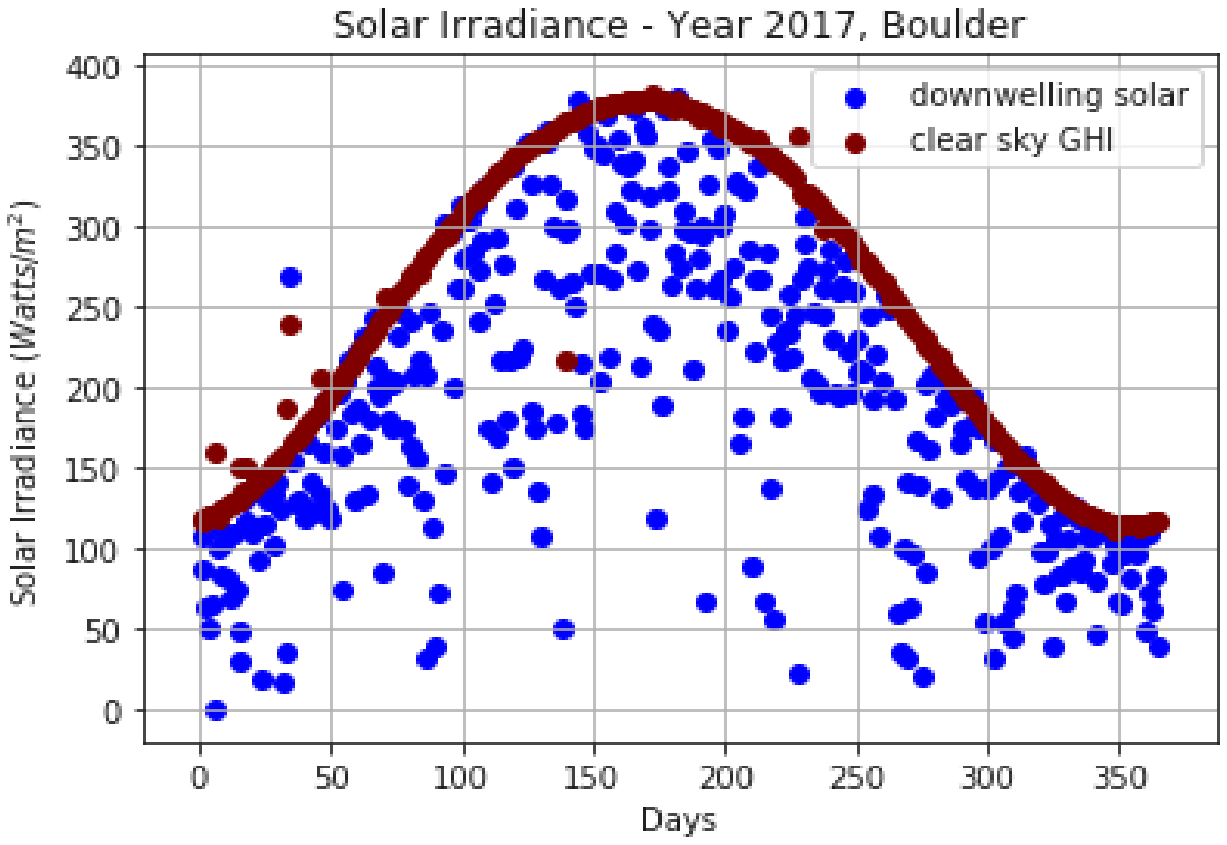

**Figure 7 Measured irradiance and clear-sky GHI**

As an interesting observation, the average daily solar radiation clear-sky GHI values (y-axis) in Figure 8 are saturating close to 400 $\text{Watts/m}^2$ for this specific location and year (Boulder, 2017). On the y-axis, the observed maximum global downwelling solar value is also below 400 $\text{Watts/m}^2$. This validates the usefulness of the clear-sky model for estimating the solar irradiance for the clear-sky days.

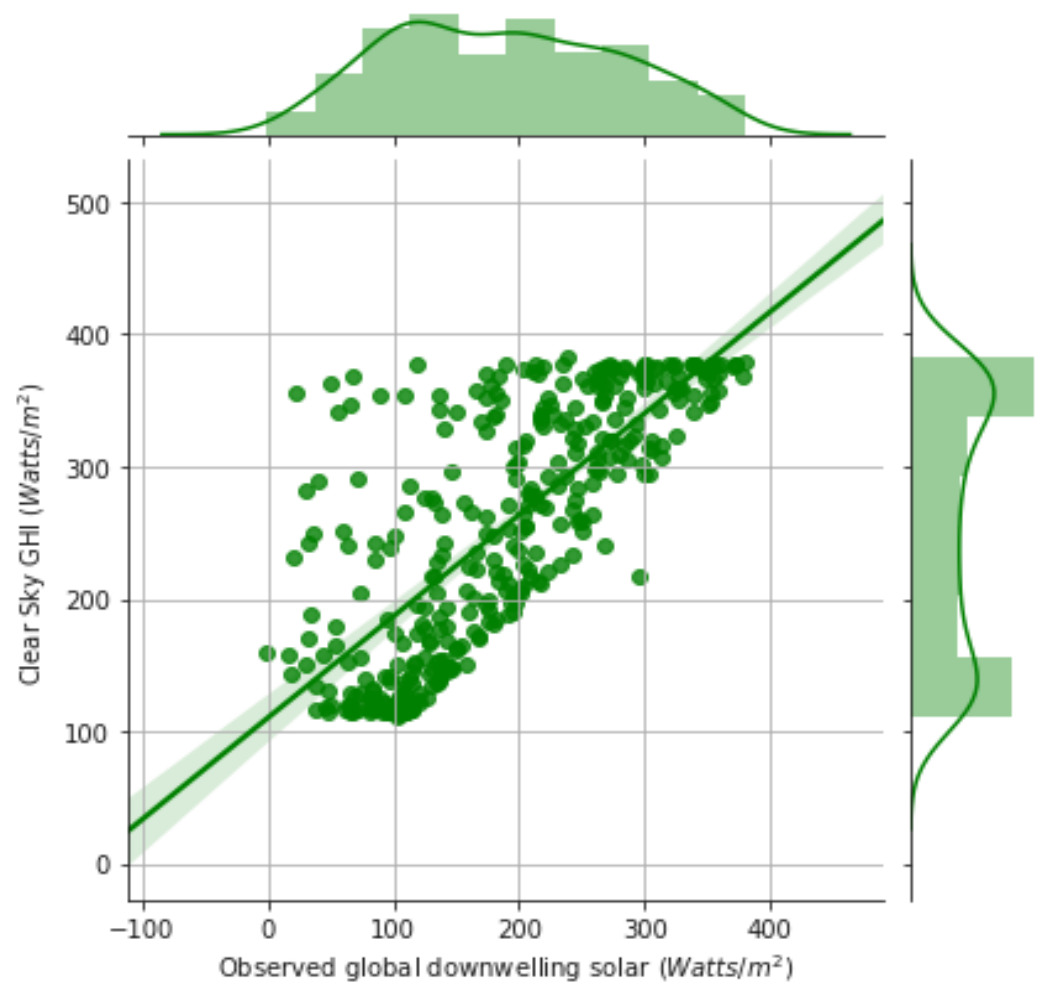

Figure 8. Correlation plot between clear-sky GHI values and measured downwelling solar irradiance

*B. Results*

The following two subsections present the results of both fixed time horizon (using RNNs) and multi-time-horizon (using both RNNs and LSTMs) approaches.

*1) Fixed-time-horizon*

The first approach generates predictions for 1-hour, 2-hour, 3-hour, and 4-hour time horizons independently using the RNN algorithm in the proposed unified architecture. Thus, there are four different models for each time horizon, and each of the seven SURFRAD sites have different models trained and tested for them. In this way, the models are temporally as well as spatially decoupled. This approach has been chosen in order to benchmark the results with the existing literature where different models were trained for different time horizons. The test loss (MSE values) over 1,000 epochs (where each epoch is one full iteration over the entire dataset), with a test set of 3,300 samples for the test year 2009, and a batch size of 100 (where batch size is a hyperparameter), for all seven sites is shown in Figure 9. This test loss is calculated as the mean squared error between the normalized predicted $K_t$ values and normalized actual $K_t$ values.

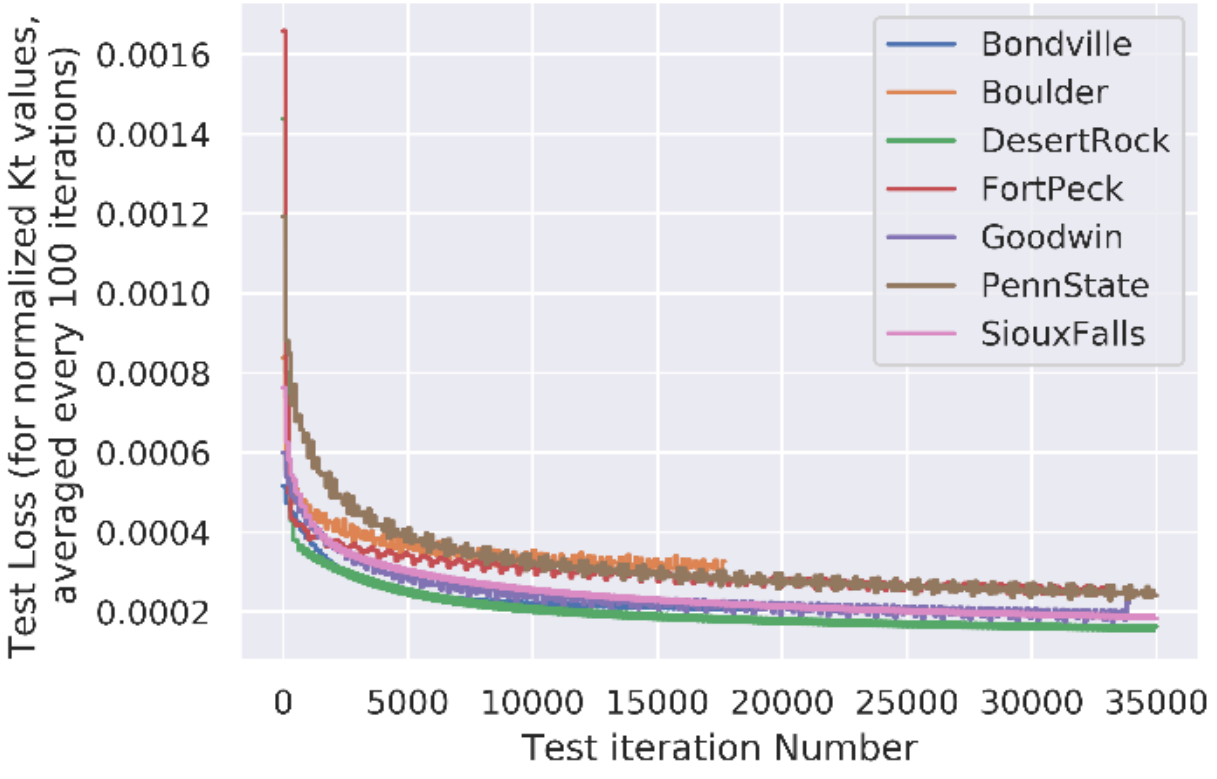

Figure 9. Test mean squared error for the SURFRAD sites

In Table 2, the denormalized RMSE values (in Watts/$m^2$) under the columns titled *RNN* are the results from our unified deep learning approach, and the RMSE values under the columns titled ML are the results reported in [43], where popular machine learning algorithms (Support Vector Machine, Random Forests, vanilla Feed-Forward, and Gradient Boosting) were implemented to compare their relative performance for short-term solar forecasting. The results in [43] are benchmarked against the persistence-of-cloudiness as well as Perez et al. [44] forecasts. Thus, this benchmarking with the "all year" category results reported in [43] intrinsically accomplishes benchmarking with the persistence models, as well. As shown in the table, our method achieves lower RMSE values for all seven sites and all four time-horizons.

**Table 2.**
**Fixed-time horizon prediction results: RMSE values for GHI predictions (Watts/$m^2$) using the proposed method (RNN) and results from [43] (ML)**

| Year 2009 | Bondville | | Boulder | | Desert Rock | | Fort Peck | | Goodwin Creek | | Penn State | | Sioux Falls | |
|---|---|---|---|---|---|---|---|---|---|---|---|---|---|---|
| F.H. | RNN | ML | RNN | ML | RNN | ML | RNN | ML | RNN | ML | RNN | ML | RNN | ML |
| 1-hour | 16.8 | 62 | 17 | 74 | 41.7 | 52 | 21.2 | 56 | 24.8 | 71 | 8.64 | 67 | 27.2 | 52 |
| 2-hour | 20.73 | 98 | 20.7 | 108 | 57.23 | 72 | 29.7 | 81 | 25.2 | 103 | 10.5 | 97 | 32.1 | 81 |
| 3-hour | 18.78 | 116 | 21.2 | 123 | 60.54 | 83 | 25.5 | 94 | 26.9 | 125 | 11.8 | 114 | 30.6 | 96 |
| 4-hour | 17.98 | 121 | 22.9 | 125 | 49.71 | 82 | 29.4 | 93 | 22 | 120 | 10.7 | 117 | 35.3 | 103 |
| Mean RMSE | 18.57 | 99.25 | 20.45 | 107.5 | 52.29 | 72.25 | 26.45 | 81 | 24.73 | 104.8 | 10.41 | 98.75 | 31.3 | 83 |

2) *Multi-time-horizon*

This approach incorporates the temporal coupling in a single model. The unified architecture predicts for all four forecasting time horizons (1-hour, 2-hour, 3-hour, and 4-hour) in parallel. For each of the SURFRAD sites, there is only one model trained that makes predictions for all four time-horizons. This method is thus the multi-time-horizon implementation of the proposed unified deep-learning-based architecture. The multi-time-horizon prediction capability is unique to the proposed deep-learning-based unified architecture, because none of the methods discussed in the literature have been shown to have this capability. The performances of the RNN and LSTM based algorithms are compared for this approach. The RMSE values for all seven sites, for the test years (2009, 2015, 2016, 2017), with RNN as well as LSTM, are listed in Table 3. For the overall performance comparison between RNN and LSTM based predictive models, the mean of the RMSEs for the four different time-horizon predictions is calculated. Based on the RMSE values for RNN and LSTM, there is no significant difference between the performance of RNN and LSTM for the short-term solar forecasting problem studied in this work.

**Table 3.**
**Multi-time-horizon prediction results: RMSE values for $K_t$[3] using RNN and LSTM**

| Year | Forecast Horizon | Bondville | | Boulder | | Desert Rock | | Fort Peck | | Goodwin Creek | | Penn State | | Sioux Falls | |
|---|---|---|---|---|---|---|---|---|---|---|---|---|---|---|---|
| | | RNN | LSTM | RNN | LSTM | RNN | LSTM | RNN | LSTM | RNN | LSTM | RNN | LSTM | RNN | LSTM |
| 2009 | 1-hour | 0.706 | 0.653 | 0.593 | 0.576 | 0.469 | 0.459 | 0.709 | 0.721 | 0.647 | 0.665 | 0.679 | 0.688 | 0.686 | 0.710 |
| | 2 -hour | 1.119 | 1.183 | 1.055 | 1.059 | 0.984 | 0.961 | 1.352 | 1.335 | 1.173 | 1.204 | 1.134 | 1.114 | 1.247 | 1.228 |
| | 3-hour | 4.524 | 4.295 | 4.010 | 4.446 | 4.527 | 4.709 | 6.311 | 6.178 | 4.586 | 4.580 | 3.161 | 3.117 | 5.061 | 4.992 |
| | 4-hour | 52.13 | 43.648 | 49.453 | 58.883 | 153.27 | 126.87 | 69.987 | 77.993 | 58.984 | 54.48 | 28.396 | 24.696 | 73.409 | 77.16 |
| | Mean RMSE | 14.62 | 12.45 | 13.78 | 16.24 | 39.81 | 33.25 | 19.59 | 21.56 | 16.35 | 15.23 | 8.34 | 7.40 | 20.10 | 21.02 |
| 2015 | 1-hour | 0.717 | 0.665 | 0.559 | 0.563 | 0.437 | 0.435 | 0.689 | 0.692 | 0.643 | 0.663 | 0.734 | 0.739 | 0.680 | 0.716 |
| | 2-hour | 1.166 | 1.195 | 1.026 | 1.041 | 0.952 | 0.929 | 1.302 | 1.273 | 1.190 | 1.229 | 1.180 | 1.179 | 1.252 | 1.246 |
| | 3-hour | 4.719 | 4.536 | 4.006 | 3.996 | 4.268 | 4.446 | 5.845 | 5.709 | 4.764 | 4.758 | 3.118 | 3.189 | 5.216 | 5.128 |
| | 4-hour | 57.84 | 53.194 | 49.349 | 52.093 | 85.952 | 74.042 | 57.687 | 61.246 | 35.814 | 40.10 | 25.796 | 23.992 | 50.823 | 50.83 |
| | Mean RMSE | 16.11 | 14.90 | 13.74 | 14.42 | 22.90 | 19.96 | 16.38 | 17.23 | 10.60 | 11.69 | 7.71 | 7.26 | 14.49 | 14.48 |
| 2016 | 1-hour | 0.731 | 0.685 | 0.593 | 0.596 | 0.456 | 0.452 | 0.726 | 0.720 | 0.669 | 0.676 | 0.738 | 0.746 | 0.723 | 0.749 |
| | 2-hour | 1.184 | 1.236 | 1.079 | 1.107 | 0.979 | 0.965 | 1.372 | 1.335 | 1.204 | 1.229 | 1.187 | 1.169 | 1.331 | 1.315 |
| | 3-hour | 4.639 | 4.407 | 4.227 | 4.193 | 4.685 | 4.777 | 6.406 | 6.147 | 4.760 | 4.708 | 3.174 | 3.118 | 5.781 | 5.710 |
| | 4-hour | 69.84 | 57.910 | 89.990 | 95.663 | 90.553 | 81.190 | 73.523 | 77.903 | 45.026 | 48.15 | 21.994 | 19.696 | 56.113 | 56.87 |
| | Mean RMSE | 19.10 | 16.06 | 23.97 | 25.39 | 24.17 | 21.85 | 20.51 | 21.53 | 12.92 | 13.69 | 6.77 | 6.182 | 15.99 | 16.16 |
| 2017 | 1-hour | 0.744 | 0.696 | 0.593 | 0.595 | 0.444 | 0.439 | 0.726 | 0.723 | 0.649 | 0.665 | 0.711 | 0.712 | 0.721 | 0.749 |
| | 2-hour | 1.190 | 1.232 | 1.059 | 1.081 | 0.949 | 0.930 | 1.368 | 1.342 | 1.179 | 1.223 | 1.149 | 1.147 | 1.323 | 1.308 |
| | 3-hour | 4.577 | 4.341 | 4.010 | 4.025 | 4.319 | 4.449 | 6.245 | 6.108 | 4.593 | 4.606 | 3.211 | 3.290 | 5.715 | 5.666 |
| | 4-hour | 81.43 | 65.839 | 49.453 | 53.488 | 44.589 | 40.318 | 114.072 | 124.41 | 32.431 | 35.64 | 29.307 | 25.779 | 59.761 | 60.74 |
| | Mean RMSE | 21.99 | 18.03 | 13.78 | 14.80 | 12.58 | 11.53 | 30.60 | 33.15 | 9.71 | 10.53 | 8.60 | 7.73 | 16.88 | 17.11 |

[3] The clear-sky index $K_t$ is defined as the ratio of the measured irradiance to the modeled clear-sky irradiance at ground level; it is a dimensionless parameter.

### *C. Results Discussion*

As described in the previous section, fixed-time-horizon predictions using our proposed approach results in a lower RMSE compared to the best-performing ML approach [43]. However, the performance of our model varies widely across different sites. The lowest RMSE is 8.64, which is obtained for Penn State site; whereas Desert Rock site with 41.7 RMSE has the worst performance. The prediction performance of neural-network-based models essentially depends on its ability to extract and represent the relationships between various input variables and the target output(s) where, the relationships can be non-linear. The input features used for training the proposed solar irradiance forecasting model include: previous time-steps' wind speed, relative humidity, wind direction, and atmospheric pressure. Cloud cover is a function of both moisture and temperature, and the shape of the clouds is affected by the wind speed to some extent, depending on the altitude of the cloud cover. We therefore hypothesize that, when trained with a reasonable length of time-series data (which is two years in our case study), our proposed RNN/LSTM-based model is capable of predicting the irradiance with better accuracies at a hilly location like Boulder by intrinsically taking cloud cover's impact into account. This powerful, non-linear, time-varying function learning capacity of RNNs and LSTMs in our opinion, leads to much better performance compared to ML methods discussed in [43].

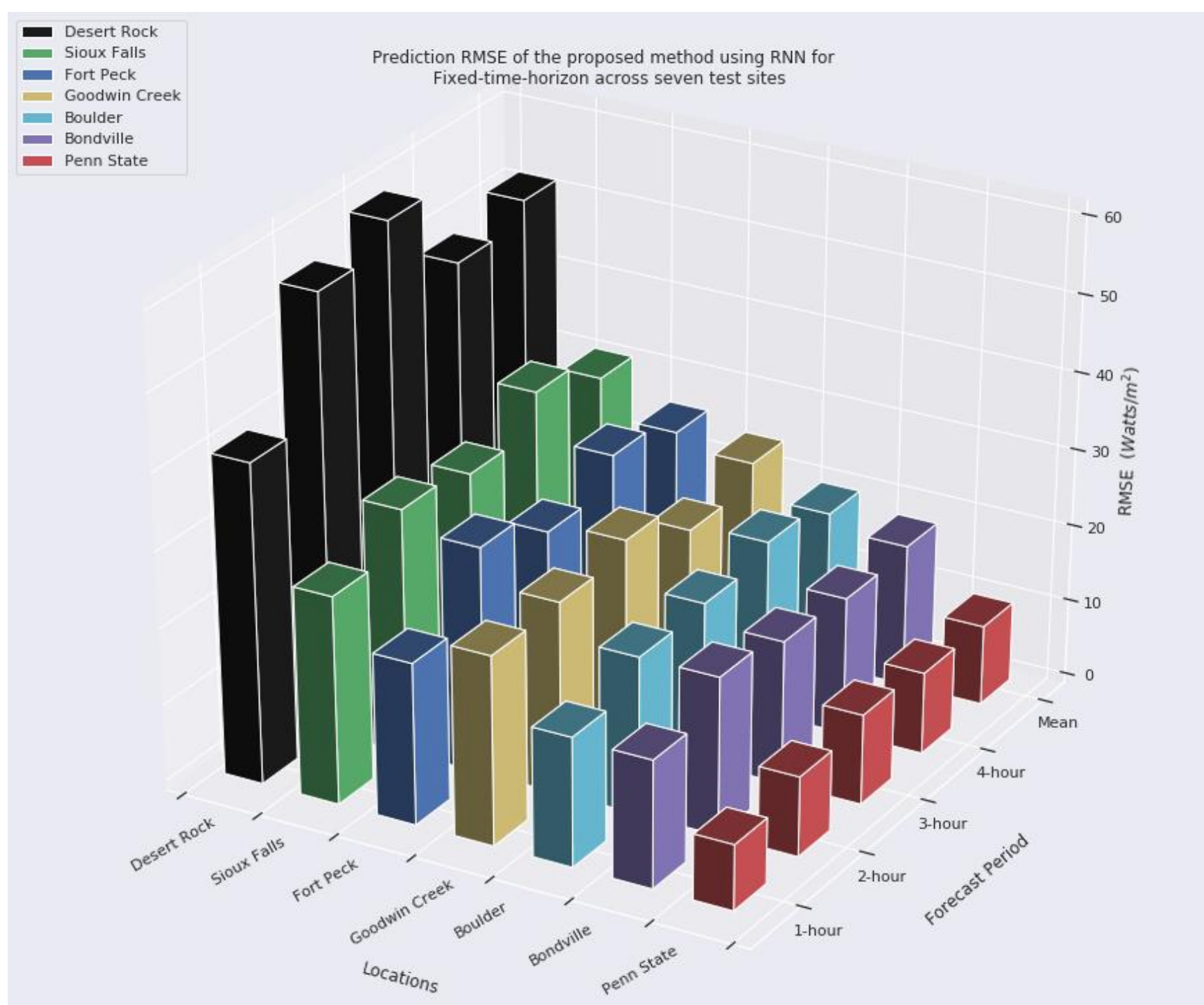

**Figure 10 Prediction error comparison between seven sites and four time-horizons for the proposed RNN-based fixed-time-horizon architecture, year 2009**

The performance of data-driven methods dependend on the quality of input data being used to train the models. Although in the data-preprocessing step, we have removed the outliers from the dataset, there lies a potential for some systematic bias in the training set, which is likely the reason for lower performance of the model for some specific sites like Desert Rock. The three-dimensional bar plot shown in Figure 10 depicts the performance of our proposed RNN-based method over different site and different fixed time-horizons. The test sites are ordered based on the mean RMSE across the time-horizons. The error magnitudes for a given site over different time-horizon are close to each other which signifies that the predictions errors are systematically related to the relationships between the input features and the forecast target. While there is no clear trend in the variation of the prediction errors in relation to the forecast period, it can be observed that the prediction error for the shortest time-horizon (1-hour) is the lowest compared to the errors for the three other long prediction time-horizons, given a site.

The *average mean RMSE* provides a useful way to measure the overall performance of the model when tested for all the seven geographically diverse locations. The average of mean RMSE for our unified deep learning approach is 26.31 Watts/m$^2$, and it is 92.36 Watts/m$^2$ for the ML-based approach. Therefore, our model shows a performance improvement of 71.5% [calculation: (92.36 – 26. 31) /92.36 ] w.r.t to the ML-based approach. Figure 11 compares the errors in forecasting between our proposed RNN-based approach and the best-performing ML approach analyzed in [43] with a bubble plot. The green bubbles are contained in the orange bubble for all the site, depicting that our proposed method outperforms the best-performing ML approach reported in [47] for all the sites.

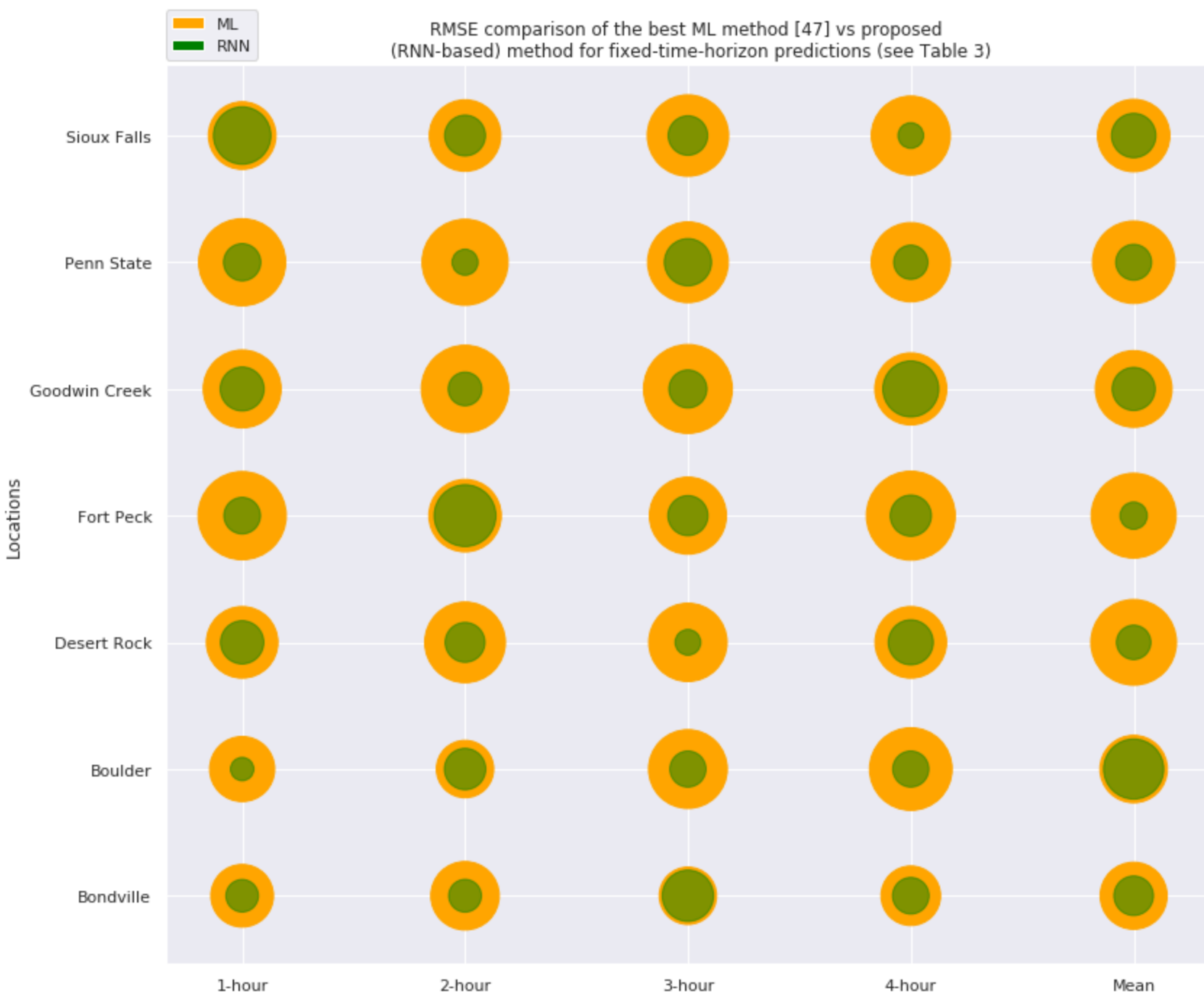

**Figure 11 Error comparison between RNN and ML algorithms for fixed-time-horizon architecture, year 2009**

For the multi-time-horizon methodology, the comparison between the proposed RNN-based and LSTM-based unified architectures shows that the performance of both the algorithms is comparable, as inferred from Figure 12. Both the algorithms show similar variation in the performance across different geographical locations. The error is minimum for Penn State and is the maximum for Fort Peck location. Recurrent neural networks with LSTMs are more stable in learning long-term dependencies, and their performance is likely to improve considerably over vanilla RNNs when a longer forecasting horizon (e.g., day-ahead) is considered.

In recent works, the use of deep learning in the prediction of solar irradiance and PV production has been presented. PV production forecasting using deep learning is tackled in [31]. However, the algorithm's design and the framework take the NWP forecasts as input, making it ineffective for intra-hour power generation prediction applications. Also, the NWP input, being a forecast itself, has a certain amount of error associated, which then gets propagated to the PV production forecasting algorithm, leading to a lower overall performance. The work presented in [32] outputs aggregated daily PV production forecasts, limiting the scope to daily forecasting. The problem formulation in [33] doesn't consider exogenous variable input. The training and testing period of the model is a total of four days, and the time resolution is 10-millisecond. Sky images and historical power output values are used in [45] for predicting the power output of a PV panel on a minute scale. Day-ahead PV power output forecasting using deep learning is presented in [46].

In our proposed deep learning algorithm based unified architecture, hourly solar irradiance is predicted based only on past time-series data consisting of exogenous atmospheric variables. Furthermore, it is capable of making predictions for intra-day multi-time-horizon using a single model, which is a novel feature, exclusive to our proposed unified architecture. We trained our models using the data for the year 2010 and 2011 and tested for four different test years (2009, 2015, 2016, 2017), and seven climatically diverse geographical locations scattered throughout the United States, providing a robust ground for the assessment of the proposed framework with the case study.

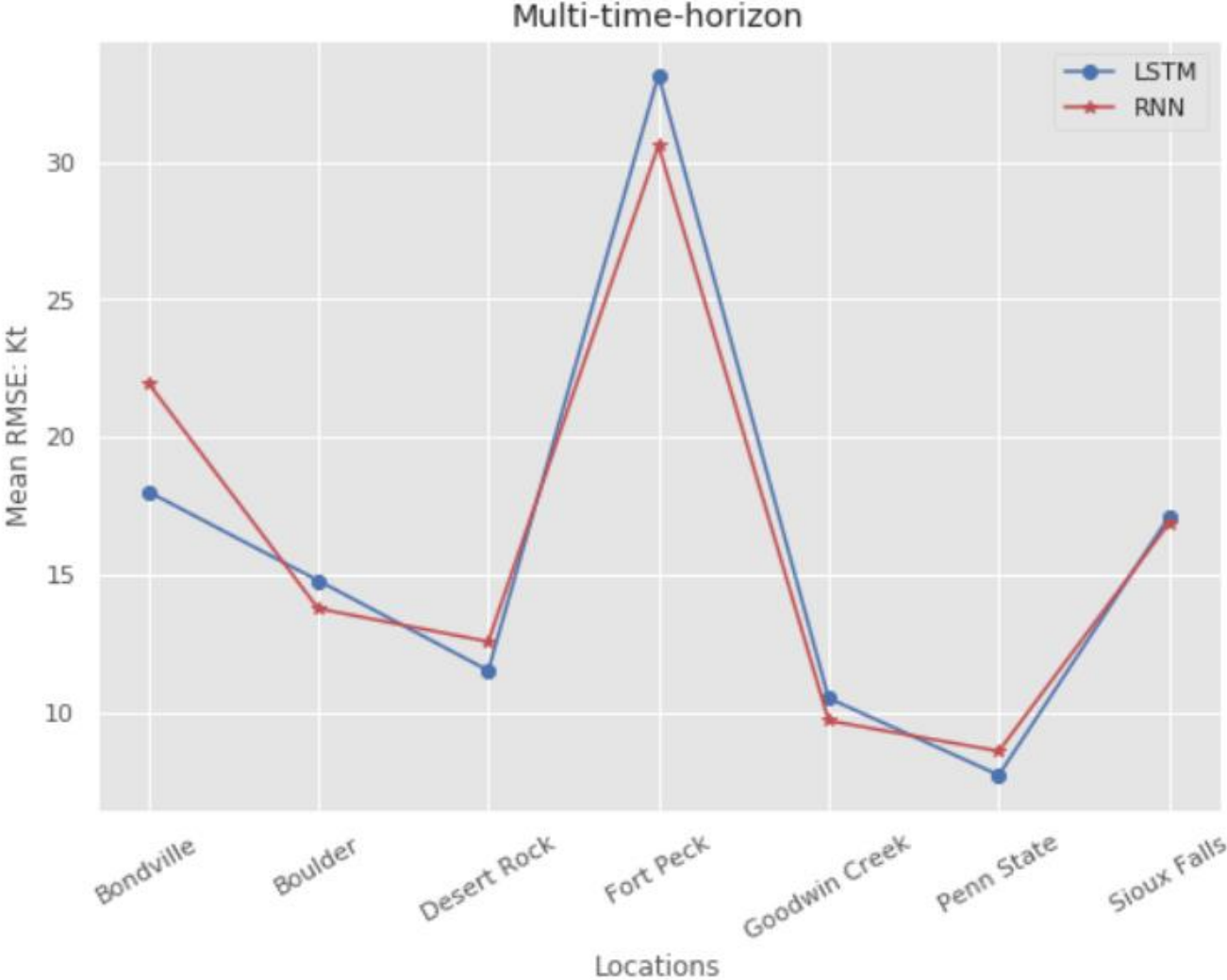

**Figure 12 Error comparison between RNN and LSTM algorithms for multi-time-horizon architecture, year 2017**

## V. SUMMARY AND OUTLOOK

Given the increased demand and generation uncertainties in the power system, the optimization of the operational efficiency and resilience of the evolving smart grids is a complex decision-making problem. In this context, improved short-term solar forecasting is of paramount importance. High-accuracy short-term solar forecasting can minimize one avenue of uncertainty: renewable forecasting errors. This can make a considerably positive impact on the operational efficiency of the grid.

Depending on the horizon of interest and application area of the predictions, a number of applications and methods have been proposed in the literature for solar irradiance forecasting. These methods have had acceptable levels of predictive success for the given targeted time horizon. However, the methods in the existing literature are relevant for a specific time horizon and cannot be employed to make predictions for different time horizons with the same model. Moreover, these methods do not implicitly or explicitly capture the impact of various other meteorological variables on solar irradiance by considering highly complex nonlinear relationships with them at any given time step. ANNs are capable of doing so. In the case of ANN methods, vanilla feed-forward neural networks do not account for the temporal relationship between various samples of the time-series data. RNNs and LSTM networks do take this temporal aspect into account. Thus, implementation of RNNs and LSTM networks for solar irradiance forecasting is an effective way to model and predict this high-frequency, non-stationary meteorological variable.

The traditional limitation of RNNs in being difficult to train has already been overcome by the recent advances in parallelism, network architectures, GPUs, and performance-optimization techniques. Thus, RNNs present promising applicability in addressing predictive analytics problems. It is one of the attributes of AI (especially DNNs) that with the increasing amount of data, the performance of the algorithms keeps increasing. Essentially, DNNs thrive on data. As the cost of sensors is decreasing rapidly, the plethora of meteorological measurement data is getting captured and made available. Thus, the performance of DNNs for predicting solar irradiance will keep increasing as the training data increases.

The proposed novel unified deep learning architecture harnesses the power of basic RNNs (and a special type of RNN called LSTM) to model a high-fidelity solar forecaster. The proposed architecture has the potential to be implemented as a unified forecasting model spanning a large portion of the temporal spectrum (sub-hourly to hourly) for short-term solar forecasting. It is also capable of taking real-time data measurements as input to produce multi-time-horizon predictions. Moreover, as demonstrated in Table 2, the proposed architecture outperforms traditional ML methods by attaining higher accuracies for all seven locations and all four horizons. The forward inference time of this proposed algorithm is low, which makes it capable of robust, near-real-time solar forecasting. We also propose the framework to extend the present unified architecture to incorporate intra-hour predictions.

The capability to predict for multi-time-horizons makes the proposed deep learning based method relevant for real-time industrial applications. The real-time measurements of the meteorological variables can be fed to the RNN and LSTM. With the lower forward inference time of the proposed forecaster, predictions can be made for multiple time scales in near-real-time. The proposed method is implemented using PyTorch, a deep-learning library, and the code for both RNN and LSTM implementations and additional information are made available on this site: https://github.com/sakshi-mishra/LSTM_Solar_Forecasting and https://github.com/sakshi-mishra/solar-forecasting-RNN.

Several ways to further improve the performance of the proposed method include architectural changes (e.g., different activation functions) and hyper-parameter tuning. The future directions for further research in this domain include exploring other variations of RNNs (e.g., Gated Recurrent Unit), accounting for uncertainty (probabilistic forecasting), and modeling the PV plant output prediction along with the solar irradiance prediction. Additionally, the relative importance of the various input variables (in the context of predicting solar irradiance) included in this study can be studied further to reduce the number of input features required for the implementation of this model.

One limitation of this approach is spatial-decoupling. Since the supervised training approach relies on the data from the given geographic location to learn the relationships between the various meteorological inputs and the GHI, it is difficult to transport/reuse this model for a climatically-different geographical location. Thus, the model will need to be trained with a location-specific dataset for using it in various geographical locations. Therefore, the algorithm itself is useful for various locations, but the model needs to be re-trained for different locations. Due to the high capacity of deep recurrent neural networks (DRNNs), the DRNN-based prediction models are prone to over-fitting. Neural networks, in general, offer a black-box approach to forecasting and lack explainability, especially when the predictions are very noisy. One useful direction of future work is in modeling the uncertainty in the prediction estimate to better capture the variability in the predictions from the DRNNs.